\documentclass[runningheads]{llncs}

\usepackage{hyperref}
\usepackage{url}
\usepackage{graphicx}
\usepackage{subcaption}
\usepackage{comment}
\usepackage{booktabs}
\usepackage{multirow}
\usepackage{amssymb}
\usepackage{adjustbox}
\newcommand{\STAB}[1]{\begin{tabular}{@{}c@{}}#1\end{tabular}}

\title{ImagiFilter: A resource to enable the semi-automatic mining of images at scale.}
\titlerunning{ImagiFilter: enabling the semi-automatic mining of images at scale.}

\author{Houda Alberts\inst{1}\thanks{$\quad$Work conducted in the University of Amsterdam as part of her MSc. research.} 
\and
Iacer Calixto\inst{2,3} 
}
\authorrunning{H. Alberts and I. Calixto}
\institute{Deloitte NL \and
ILLC, University of Amsterdam \and CIMS, New York University\\
\email{houda.alberts@gmail.com}, \email{iacer.calixto@nyu.edu} }

\begin{document}

\maketitle

\begin{abstract}Datasets (semi-)automatically collected from the web can easily scale to millions of entries, but a dataset's usefulness is directly related to how clean and high-quality its examples are. In this paper, we describe and publicly release an image dataset along with pretrained models designed to (semi-)automatically filter out undesirable images from very large image collections, possibly obtained from the web.
Our dataset focusses on photographic and/or natural images, a very common use-case in computer vision research. We provide annotations for coarse prediction, i.e. photographic vs. non-photographic, and smaller fine-grained prediction tasks where we further break down the non-photographic class into five classes: maps, drawings, graphs, icons, and sketches.
Results on held out validation data show that a model architecture with reduced memory footprint achieves over $96\%$ accuracy on coarse-prediction. Our best model achieves $88\%$ accuracy on the hardest fine-grained classification task available. Dataset and pretrained models are available at: \url{https://github.com/houda96/imagi-filter}.
\end{abstract}

\section{Introduction}
\label{sec:intro}

Datasets collected (semi-)automatically from the web can scale to millions of examples, allowing researchers and practitioners to deal with amounts of data that were unimaginable just a decade before.
However, the usefulness of a dataset depends on having clean and high-quality examples.
More specifically, in computer vision there is an obvious need for large labelled datasets with high-quality photographic image collections, e.g. ImageNet~\cite{imagenet_cvpr09,ILSVRC15}.
However, collecting, annotating, and curating such datasets is an expensive and time-consuming effort.

\begin{figure*}[t!]
\begin{subfigure}[t]{0.15\textwidth}
  \centering
  \includegraphics[width=\textwidth]{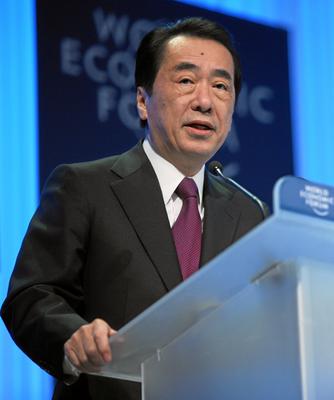}
  \caption{Positive class.}
\end{subfigure}%
\hfill
\begin{subfigure}[t]{.12\textwidth}
  \centering
  \includegraphics[width=\textwidth]{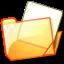}
  \caption{\textit{Icon}.}
\end{subfigure}%
\hfill
\begin{subfigure}[t]{.15\textwidth}
  \centering
  \includegraphics[width=\textwidth]{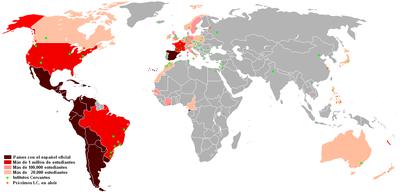}
  \caption{\textit{Map}.}
\end{subfigure}%
\hfill
\begin{subfigure}[t]{0.18\textwidth}
  \centering
  \includegraphics[width=\textwidth]{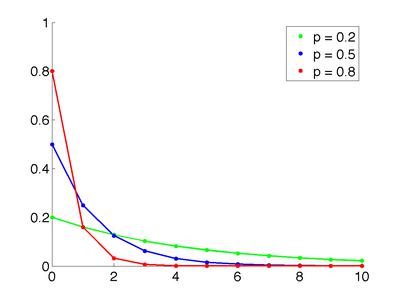}
  \caption{\textit{Graph}.}
\end{subfigure}%
\hfill
\begin{subfigure}[t]{.11\textwidth}
  \centering
  \includegraphics[width=\textwidth]{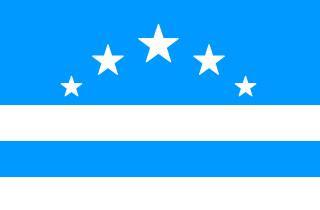}
  \caption{\textit{Flag}. }
\end{subfigure}
\hfill
\begin{subfigure}[t]{.12\textwidth}
  \centering
  \includegraphics[width=\textwidth]{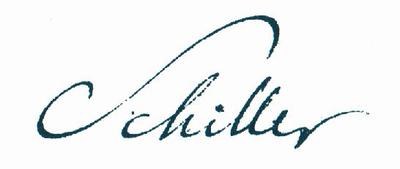}
  \caption{\textit{Sketch}.}
\end{subfigure}
\hfill
\begin{subfigure}[t]{.14\textwidth}
  \centering
  \includegraphics[width=\textwidth]{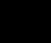}
  \caption{\textit{Other}.}
\end{subfigure}
\caption{Examples of images labeled positive \textit{(a)} and negative \textit{(b)--(g)}, as well as images' fine-grained negative labels.}
\label{fig:negatives}
\end{figure*}

In this paper, we describe and publicly release the \textit{ImagiFilter} dataset and pretrained models to facilitate the process of (semi-)automatically collecting high-quality image datasets.
We also release pretrained convolutional neural network (CNN) models trained to filter out different sets of \textit{undesirable images} from possibly very large image collections.
We denote undesirable images as those that are not natural or photographic, including categories such as maps, sketches, and drawings.
Our models are very simple and can be directly applied to any image collection pipeline to (semi-)automatically tag and/or filter out images. 

We conduct a set of experiments where we investigate how consistently our proposed models based on different CNN architectures \textit{(i)} predict images as photographic or not, i.e. \textit{coarse prediction}, and \textit{(ii)} predict specific negative classes individually, i.e. \textit{fine-grained prediction}.
We evaluate different CNN architectures pretrained on ImageNet~\cite{ILSVRC15}, i.e. a 19-layer VGG network~\cite{Simonyan15VGGNet} and a 152-layer Residual Network~\cite{He2015ResNet}, and also a 5-layer LeNet~\cite{Lecun98LeNet} trained from scratch.

We find that even simple architectures such as the 5-layer LeNet network trained from scratch on our coarse prediction task already achieves over $96\%$ accuracy predicting images as photographic or not.
Its negligible memory footprint makes it a strong candidate in case the $\sim90\%$ accuracy is acceptable in the actual image collection pipeline they are to be used, especially if the number of images to be processed in parallel is very large and/or the hardware infrastructure is restricted, e.g. one is deploying the model on mobile phones.


Predicting fine-grained non-photographic classes is harder due to the smaller number of labelled images and unbalanced number of examples in positive and negative classes, and in such cases newer CNN architectures pretrained on ImageNet are necessary for good results.
Our best model on different fine-grained classification tasks (explained in detail in Section~\ref{sec:approach}) is a ResNet-152 pretrained on ImageNet and further fine-tuned on each task, and achieves validation accuracies between $87\%$ and $98\%$.
We discuss all results in detail in Section~\ref{sec:experiments}.

Our main contributions are the public release of:

\begin{itemize}
    \item the \textit{ImagiFilter dataset}, consisting of $6,000$ images annotated with balanced binary labels, i.e. photographic and non-photographic, and additional fine-grained labels for non-photographic images, making it possible to train models to predict fine-grained negative classes (see Figure~\ref{fig:negatives} for example data).
    \item code to readily use ImagiFilter to train coarse- and fine-grained image filtering models, as well as pretrained models with example code to use them.
\end{itemize}


\section{Related work}
\label{sec:related}

Over the years, many high-quality photographic image datasets have been publicly released to the research community, such as ImageNet~\cite{imagenet_cvpr09}, Flickr-30k~\cite{flickr30k}, CIFAR-10~\cite{krizhevsky2009learning}, Fashion MNIST \cite{xiao2017fashion}, MSCOCO~\cite{lin2014microsoft}, among many others.
These are all very valuable resources, and each required considerable validation and annotation effort, especially larger ones such as ImageNet.
Usually, the data collection pipeline would involve automatically obtaining a large number of possibly high-quality images from the web, for which human-verified labels and annotations would be most likely crowd-sourced using tools such as Amazon Mechanical Turk (AMT)\footnote{\url{https://www.mturk.com/}} or a similar tool.

Existing work that is closest to what we put forward in this work propose models to automatically detect and filter out pornographic images from large image collections~\cite{Deselaersetal2008_bagofvisualwords,moustafa2015applying,Fralenko2018automatic}, or to automatically detect pornographic and/or inappropriate content from videos~\cite{Caetanoetal2014BossaNova}.

To the best of our knowledge, we are the first to publicly release a dataset and models to filter out non-photographic images from a large image collection.
Possibly, this is partially because the problem we address is simple, however we believe it might still be useful for the community.
The ImagiFilter dataset we describe in this work can be used in \textit{(i)} a semi-automatic fashion to select candidates among a very large set of images to be sent to be manually annotated by crowd-workers,
or in principle could \textit{(ii)} directly replace AMT and automatically filter out noisy images from a set of images downloaded from the internet.


\section{Approach}
\label{sec:approach}

We design the ImagiFilter dataset to address a real problem we encountered:
to ensure the quality of images collected (semi-)automatically from the web, and to be able to filter out undesirable images and ensure a high-quality image dataset with minimal manual annotation/interference.
Although the actual image collection pipeline we applied ImagiFilter to has influenced some decisions on how we built the ImagiFilter dataset, we tried to make it as general and as useful as possible to other researchers and practitioners facing similar issues.\footnote{The dataset we built using ImagiFilter is not within the scope of this paper, therefore we do not discuss it any further.} In Table \ref{tab:stats_imagi} we show dataset statistics.

\begin{table}[t!]
    \centering
    \begin{tabular}{clr}
    \toprule
    && \bf \# images \\
    \midrule
    & \bf Positive class & $3,000$\\
    \midrule
    \multirow{6}{*}{\STAB{\rotatebox[origin=c]{90}{\bf Negative class}}}
    & Hand drawings/sketches & $157$ \\
    & Flags & $372$ \\
    & Graphs/rendered images & $425$  \\
    & Icons & $810$ \\
    & Maps & $1,042$  \\
    & Other & $206$ \\
    \cmidrule{2-3}
    & {\bf Subtotal (\# unique)} & $3,012$ ($3,000$)\\
    \midrule
    & {\bf Total} & $6,000$\\
    \bottomrule
    \end{tabular}
    \caption{ImagiFilter dataset statistics.}
    \label{tab:stats_imagi}
\end{table}

\paragraph{Data collection}
BabelNet~\cite{NavigliPonzetto:12aij} is a large multilingual knowledge graph that integrates many datasets and knowledge bases, including multilingual Wikipedias and WordNets~\cite{Miller1995WordNet}.
Nodes in the BabelNet graph are referred to as \textit{synsets} and can have multiple \textit{sources}, e.g. Wikipedia or WordNet, as well as images associated to them.
We first use the BabelNet API\footnote{\url{https://babelnet.org/guide}} to download images for the $1,000$ ImageNet classes\footnote{\url{http://image-net.org/challenges/LSVRC/2014/browse-synsets}
}
used in the ILSVRC image classification competition~\cite{ILSVRC15}.
We then proceed by downloading images for nodes related to these first $1,000$ nodes. 
In practice, this is done by performing a random walk on the BabelNet graph starting from one of the $1,000$ nodes and stopping at nodes which \textit{source} is either WordNet and/or Wikipedia and downloading the images available for that node.
The random walk included nodes at most six hops away from one of the original $1,000$ nodes.

After these images are collected, we manually annotate a subset of $3,000$ positive and $3,000$ negative randomly chosen images from the pool.
The annotation labels are either \textit{photographic} for the positive class, or {non-photographic} for the negative class.

\paragraph{Fine-grained negative labels} In order to make ImagiFilter more useful to the research community, we also label each image in the negative class with fine-grained labels chosen from: \textit{(i)} hand drawings and/or sketches, \textit{(ii)}  maps, \textit{(iii)} graphs and/or rendered images, \textit{(iv)}  icons, \textit{(v)} flags, and \textit{(vi)} other.
Note that the "other" label is used to tag any images that do not fall under any of the other five fine-grained negative labels and includes, among other things, images that are too dark, screenshots, 
and misplaced images.
In Figure~\ref{fig:negatives} we show examples of images labelled positive as well as labelled with each of the fine-grained negative class.
Importantly, one same (non-photographic) image can be annotated with multiple fine-grained negative labels, e.g. maps and hand-drawings.


\section{Experimental settings and results}
\label{sec:experiments}


In order to evaluate the quality of the dataset, we begin by splitting the annotated images into training and validation sets and conduct a series of experiments.
Training and validation splits include 90\%/10\% of images respectively, and are balanced with respect to positive/negative examples.

\subsection{Models}
\label{sec:models}
We use convolutional neural networks (CNNs) as our image encoders.
We compare three different CNN architectures, a VGG-19~\cite{Simonyan15VGGNet}, a ResNet-152~\cite{He2015ResNet} and a LeNet-5~\cite{Lecun98LeNet}.
The ResNet-152 and VGG-19 are pre-trained\footnote{\url{https://pytorch.org/docs/stable/torchvision/models.html}.} on ImageNet classification~\cite{ILSVRC15}, whereas the LeNet-5 CNN is trained from scratch.
We remove each model's last layer and replace it by a trained linear/fully-connected layer. 
We use the Adam optimizer with a learning rate of 1e$-$4, $\beta=(0.9, 0.999)$, $\epsilon=$~1e$-$8, no weight decay or momentum, and a batch size of 40.
Moreover, for the LeNet-5 architecture we scale images to 300x300 and we use dropout with $50\%$ probability after each convolutional block in the fine-grained tasks.

\subsection{Coarse prediction}
\label{sec:coarse_prediction}
We first train models on a \textit{binary classification} task where the goal is to classify an image as photographic or not.
Positive/negative classes are balanced across training and validation splits, i.e. training set has 2700 positive and 2700 negative examples, and validation set and 300 positive and 300 negative examples.

\subsection{Fine-grained prediction}
\label{sec:finegrained_prediction} Second, we are interested in investigating how well can different models predict fine-grained negative classes individually.
For that purpose, we train models on five additional \textit{binary classification} tasks, i.e. one per fine-grained negative class: \textit{sketch}, \textit{map}, \textit{graph}, \textit{icon}, and \textit{flag}. 
We reserve a fixed number of examples in each fine-grained class for model evaluation, and leave the remainder for training.
More specifically, classes \{\textit{map}, \textit{icon}\} each have $200$ examples for model evaluation, and all other classes \{\textit{sketch}, \textit{graph}, \textit{flag}\}
have $100$ examples each.

\subsection{Results}
\label{sec:results}

\begin{figure}
    \centering
    \includegraphics[width=0.8\linewidth]{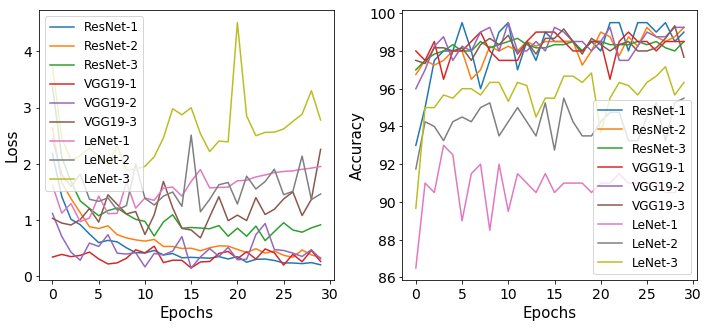}
    \caption{Validation loss and accuracy for different CNN architectures trained on coarse label prediction (i.e. binary classification).
Numbers 1, 2 and 3 denote 1000, 2000 and 3000 number of training examples per class respectively.}
    \label{fig:coarse_prediction_results}
\end{figure}

We first report results for the coarse prediction experiments.

\paragraph{Coarse prediction} In Figure~\ref{fig:coarse_prediction_results}, we show loss and accuracy curves of different CNNs VGG-19, ResNet-152, LeNet-5, trained on all available data (i.e. $3,000$ examples per class) and on subsets of the available data (i.e. $2,000$ and $1,000$ examples per class) to show the impact the amount of training data has on predictions.
In Table \ref{tab:perform} we see the performance of each model architecture over 5 different runs (mean and one standard deviation). 
On average, all models converge after around 15 epochs.

We note that the LeNet-5 trained from scratch is more sensitive to the amount of training data, as expected.
Though LeNet-5 is a simple architecture, when trained on all available examples it shows reasonable performance. 

\begin{table}[t!]
    \centering
    \begin{adjustbox}{max width=\linewidth}
    \begin{tabular}{lrrrr}
    \toprule
    \bf Model & \bf Accuracy & \bf Epoch & \bf \# params. & \bf PT? \\
    & $\mu (\sigma)$ &&  \bf trained / total \\
    \midrule
    LeNet-5 & $96.3 (0.7)$ & $15.8$ & 10M / 10M & $\times$ \\
    VGG-19 & $98.4 (0.1)$ & $16.8$ & 8194 / 140M & \checkmark \\
    ResNet-152 & $98.8 (0.4)$ & $14.2$ & 4098 / 58M & \checkmark \\
    \bottomrule
    \end{tabular}
    \end{adjustbox}
    \caption{Validation accuracy (mean and one standard deviation) per model architecture over five different runs and mean number of training epochs per model. ``PT?'' indicates if model is pre-trained on ImageNet.}
    \label{tab:perform}
\end{table}

\begin{figure}[t!]
    \centering
    \includegraphics[width=0.8\linewidth]{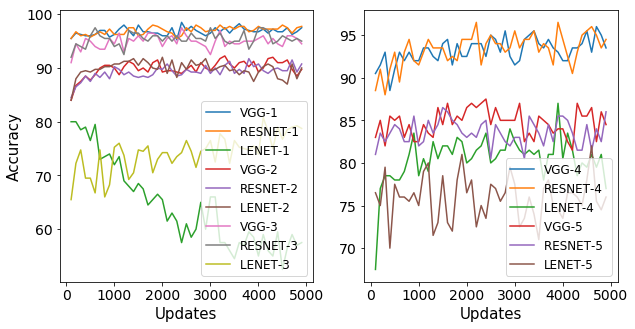}
    \caption{Validation accuracies on fine-grained label prediction over five categories for different CNN architectures, where the numbers 1--5 stand for maps, icons, graphs, flags, and sketches, in that order.}
    \label{fig:accs_fine}
\end{figure}

\paragraph{Fine-grained prediction}
In Figure~\ref{fig:accs_fine} we show accuracies
for VGG-19, ResNet-152, and LeNet-5 when predicting each of the five fine-grained negative classes \texttt{\{}\textit{sketch, map, graph, icon, flag}\texttt{\}}.
Training the LeNet-5 on fine-grained classification cause the model to mostly overfit early on.
We tried adding dropout to mitigate this issue, but still observed overfitting if categories have few training examples, e.g. sketches class has 57 training images.
Since validation accuracy still improved albeit marginally, we always report LeNet-5 results with dropout unless specified otherwise.
Models achieve best validation scores already after only 100 updates, compared to 3500--4500 updates for the other two models VGG-19 and ResNet-152; a clear indication that the LeNet-5 model is very sensitive to the amount of training data.
These results are not surprising, since it is the only of the three models not pretrained on ImageNet.

\begin{figure}[t!]
  \centering
  \includegraphics[width=0.7\textwidth]{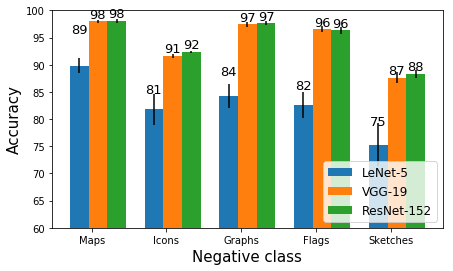}
\caption{Validation accuracy for different CNN architectures on fine-grained label prediction.}
\label{fig:finegrained_prediction_results}
\end{figure}

In Figure~\ref{fig:finegrained_prediction_results}, we note that both ResNet-152 and VGG-19 tend to perform similarly and achieve higher classification accuracy compared to the LeNet-5.
Moreover, the variance of the LeNet-5 training runs are much higher than those of the VGG-19 and ResNet-152, similarly to what we observed on the coarse prediction task but amplified, since here there are even fewer training data points.
In general, VGG-19 and ResNet-152 validation accuracies are consistently strong and between $88\%$ and $98\%$, which makes both equally good models to use to predict fine-grained negative classes.

\subsection{Error Analysis}
\paragraph{Coarse prediction.}

\begin{figure}[t!]
\begin{subfigure}[t]{0.2\textwidth}
  \centering
  \includegraphics[width=\textwidth]{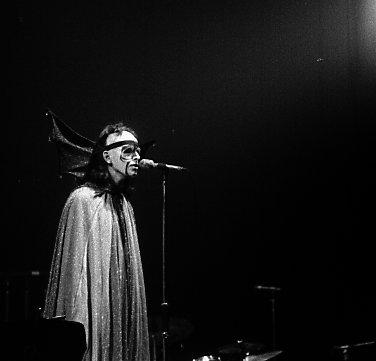}
  \caption{LeNet-5 (false negative).
  }
\end{subfigure}%
\hfill
\begin{subfigure}[t]{.2\textwidth}
  \centering
  \includegraphics[width=\textwidth]{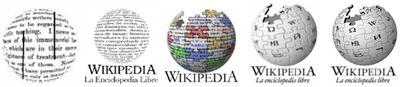}
  \caption{ResNet-152 (false positive).
  }
\end{subfigure}%
\hfill
\begin{subfigure}[t]{.2\textwidth}
  \centering
  \includegraphics[width=\textwidth]{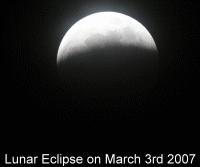}
  \caption{VGG-19 (false negative).
  }
\end{subfigure}%

\caption{Examples of mistakes, all models were trained on the complete dataset.}
\label{fig:mistakes_coarse}
\end{figure}

In Figure~\ref{fig:mistakes_coarse} we illustrate mistakes each CNN makes in the coarse prediction task.
We note that models have more difficulty predicting dark/black-and-white images (LeNet-5, VGG-19), and also show an example where logos are incorrectly predicted as positive (ResNet-152).

\begin{figure}[t!]
  \centering
  \includegraphics[width=.75\textwidth]{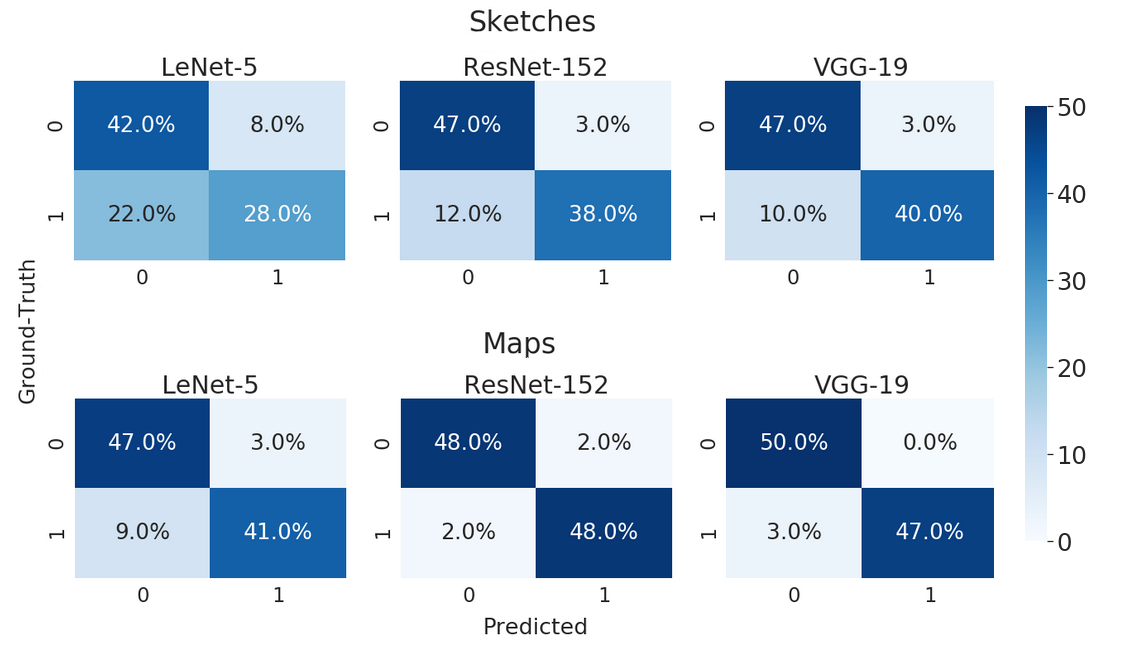}
\caption{Confusion matrices for the sketches and maps categories for each CNN architecture, and the 0 and 1 on the axes represent the non-fine-grained and fine-grained class respectively.}
\label{fig:conf_matrices}
\end{figure}

\paragraph{Fine-grained prediction} 
In Figure~\ref{fig:conf_matrices} we show confusion matrices for fine-grained prediction of maps (most represented category) and sketches (least represented category).\footnote{Other categories show similar results as maps class.} It clearly shows that the less represented fine-grained class (sketches) has much more mistakes compared to the better represented class. Each model has the tendency to make more mistakes on actual fine-grained images, which is expected due to it being less present in the dataset during learning.

\section{Final remarks}
\label{sec:remarks}
We introduce the \textit{ImagiFilter} dataset aimed to help researchers in the semi-automatic assessment of the quality of images gathered from the web.
We release CNNs trained on \textit{ImagiFilter} on a variety of classification tasks, i.e. one coarse-grained and five fine-grained tasks.
Our best model, a ResNet-152 pretrained on ImageNet and fine-tuned on our dataset, achieves validation accuracies between $88\%$ and $98\%$ across coarse- and fine-grained classification tasks.
If memory footprint is an issue, 
we also release the LeNet-5 architecture since it still achieves respectable $96\%$ validation accuracy on coarse prediction, while having 6$\times$ less parameters than the ResNet-152 CNN.



As future work, we will focus on improving our fine-grained classes.
We will collect more images for the existing negative classes that have the fewest images.
We will also add more fine-grained categories to the dataset according to the community's interests, since collecting and annotating even $\sim100$ images have shown strong predictive validation accuracy.

\section*{Acknowledgements}
This project has received funding from the European Union’s Horizon 2020 research and innovation programme under the Marie Skłodowska-Curie grant agreement No 838188.


\bibliography{refs}
\bibliographystyle{splncs04}


\end{document}